\documentclass[10pt,twocolumn,letterpaper]{article}

\usepackage{3dv}
\usepackage{times}
\usepackage{epsfig}
\usepackage{graphicx}
\usepackage{amsmath}
\usepackage{amssymb}
\usepackage{multirow}
\usepackage{color, colortbl}
\usepackage{soul}

\usepackage{verbatim}

\definecolor{Gray}{gray}{0.9}
\newcolumntype{g}{>{\columncolor{Gray}}c}

\makeatletter
\def\thickhline{%
  \noalign{\ifnum0=`}\fi\hrule \@height \thickarrayrulewidth \futurelet
   \reserved@a\@xthickhline}
\def\@xthickhline{\ifx\reserved@a\thickhline
               \vskip\doublerulesep
               \vskip-\thickarrayrulewidth
             \fi
      \ifnum0=`{\fi}}
\makeatother

\newlength{\thickarrayrulewidth}
\usepackage[pagebackref=true,breaklinks=true,colorlinks,bookmarks=false]{hyperref}

\threedvfinalcopy 

\def\threedvPaperID{408} 

\ifthreedvfinal\fi
\begin{document}

\title{VoRTX: Volumetric 3D Reconstruction With Transformers \\ for Voxelwise View Selection and Fusion}


\author{
Noah Stier
\and
Alexander Rich
\and
Pradeep Sen
\and
Tobias Höllerer
}

\date{{\tt\small\{noahstier, anrich, psen, thollerer\}@ucsb.edu}}

\affiliation{University of California, Santa Barbara}






\maketitle

\thispagestyle{empty}

\begin{abstract}

Recent volumetric 3D reconstruction methods can produce very accurate results, with plausible geometry even for unobserved surfaces. However, they face an undesirable trade-off when it comes to multi-view fusion. They can fuse all available view information by global averaging, thus losing fine detail, or they can heuristically cluster views for local fusion, thus restricting their ability to consider all views jointly. Our key insight is that greater detail can be retained without restricting view diversity by learning a view-fusion function conditioned on camera pose and image content. We propose to learn this multi-view fusion using a transformer. To this end, we introduce VoRTX,\footnote{\url{https://noahstier.github.io/vortx}} an end-to-end volumetric 3D reconstruction network using transformers for wide-baseline, multi-view feature fusion. Our model is occlusion-aware, leveraging the transformer architecture to predict an initial, projective scene geometry estimate. This estimate is used to avoid backprojecting image features through surfaces into occluded regions. We train our model on ScanNet and show that it produces better reconstructions than state-of-the-art methods. We also demonstrate generalization without any fine-tuning, outperforming the same state-of-the-art methods on two other datasets, TUM-RGBD and ICL-NUIM.

\end{abstract}

\section{Introduction}

3D reconstruction is a fundamental problem in computer vision, supporting applications such as autonomous navigation and mixed reality. In many scenarios, dense and highly detailed reconstruction is desirable. For example, it can facilitate the creation of virtual reality content by scanning real-world scenes, or the simulation of physics-based effects in augmented reality. Although active depth sensors
have been employed for this purpose \cite{dai2018scancomplete, newcombe2011kinectfusion}, they increase platform cost relative to passive cameras. It is therefore desirable to perform reconstruction using only visible-light RGB cameras, which are ubiquitous and relatively inexpensive.



\begin{figure}[t]
\begin{center}
\includegraphics[width=\linewidth]{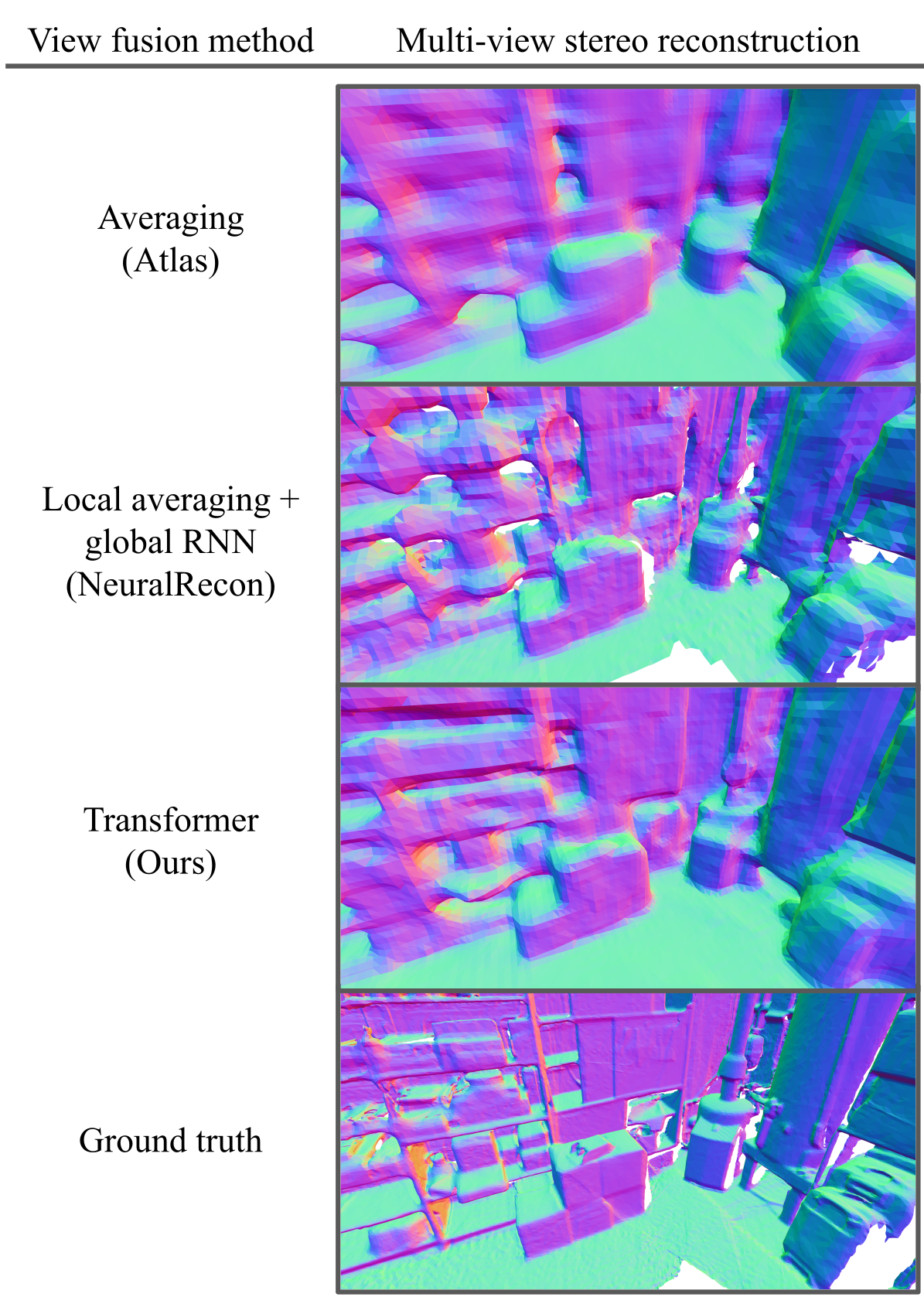}
\end{center}
\vspace{-0.3cm}
\caption{Our method fuses input view features using a transformer. We compare to Atlas \cite{murez2020atlas}, which fuses features by averaging, and NeuralRecon \cite{sun2021neuralrecon}, which fuses locally by averaging and globally by RNN. Our method produces a high level of detail, while also filling in holes due to occlusion and unobserved regions.}
\label{fig:teaser}
\vspace{-1.5em}
\end{figure}

Dense 3D reconstruction from RGB imagery traditionally consists of estimating depth for each image, and then fusing the resulting depth maps in a reprojection step. This approach, however, cannot fill holes arising from occlusions and other unobserved regions.


Recently, a number of works have addressed this by posing RGB-only 3D reconstruction as the direct prediction of a truncated signed-distance function (TSDF), using deep learning to fill in unobserved regions via learned priors \cite{murez2020atlas, sun2021neuralrecon}. These methods extract image features using a convolutional neural network (CNN), accumulate them into space by backprojecting onto a 3D grid, and then predict the TSDF volume using a 3D CNN. When a particular grid voxel is within the view frustum of multiple cameras, it is common practice to fuse the backprojected image features at that point via unweighted averaging.
However, we observe two drawbacks of directly averaging view features. 

First, when images are acquired from very different camera poses, their content may not be directly comparable. Although CNNs are capable of extracting high-level semantic features that are therefore highly view-independent, the CNN architectures commonly used in 3D reconstruction (e.g.,  U-Net \cite{ronneberger2015u} and FPN \cite{lin2017feature}) make explicit use of the activations from early CNN layers. These are understood to represent lower-level visual features, which exhibit view-independence only within a particular range of viewpoint difference. Averaging across disparate views does not take that range into consideration, and therefore loses view-dependent information. This is a known phenomenon in multi-view stereo, where typical solutions include 1) selecting views using constraints on camera pose to minimize viewpoint differences \cite{galliani2015massively, schonberger2016pixelwise}, or 2) constraining the image features to be as view-independent as possible \cite{tola2009daisy}. We hypothesize that a better solution can be obtained by learning a view fusion function, conditioned on pose and image content, that can jointly consider features from multiple views within the appropriate range of viewpoints.

Second, averaging assigns an equal weight to all input views at each voxel, including views for which a voxel is occluded. This problem is exacerbated in wide-baseline reconstruction, where occlusions are particularly prevalent. Occlusion modeling presents a chicken-and-egg problem: the scene geometry is not known until after backprojection and reconstruction; but until the scene geometry is known, backprojection cannot account for occlusions, thus projecting image features through surfaces into regions where they are irrelevant. We hypothesize that this irrelevant information acts as noise that reduces reconstruction quality.

We propose an innovation that addresses both issues. Our model, which we call VoRTX, is a deep learning-based volumetric reconstruction network using transformers \cite{vaswani2017attention} to model dependencies across diverse viewpoints. The transformers use self-attention to perform soft grouping of views that are mutually relevant, and they can learn to fuse within vs.\ across groups in different feature spaces.

Transformers also provide a natural mechanism for occlusion-awareness, since the attention to each input view varies as a function of 3D location. The view aggregation can therefore be supervised to encourage reduced attention to input images in regions where their view is occluded. One possibility is to supervise the view aggregation using ground-truth visibility. However, we argue in Sec.~\ref{section:projocc} that \textit{projective occupancy} is preferable for our problem setting, because it is an easier target that more closely describes the desired spatial distribution of image features during backprojection.
Our main contributions are as follows:
\begin{enumerate}
    \item We introduce a new method of fusing multi-view image features, using a transformer to perform data-dependent fusion at each spatial location.
    \item We propose the projective occupancy as an occlusion-aware reconstruction target for deep volumetric MVS, and we show that it yields improved results over unsupervised or visibility-supervised reconstruction.
    

\end{enumerate}

We show that VoRTX surpasses state-of-the-art reconstruction results when compared with several baseline methods, on multiple datasets.

\section{Related work}

\noindent \textbf{Image feature fusion in MVS: }
Fusing measurements from multiple views is a crucial step in MVS. Typically, image patches are fused into a cost volume using a stereo-matching cost function, which operates on raw image intensity \cite{galliani2015massively, hou2019multi, schonberger2016pixelwise, wang2018mvdepthnet} or CNN-extracted image features \cite{duzceker2021deepvideomvs, yao2018mvsnet}. Some methods \cite{huang2018deepmvs, im2019dpsnet} instead concatenate image features in the channel dimension, and use a CNN to reduce them into a cost volume. These techniques are effective when the input views are acquired closely enough in pose space to maintain similar scene appearance, while still providing enough parallax for stereopsis. 

Atlas \cite{murez2020atlas} proposes the use of a single feature volume, bypassing depth prediction and posing 3D reconstruction as the direct prediction of a TSDF volume. This is an effective way to consider all input images jointly, and it also provides a framework for learning to reconstruct unobserved scene regions via 3D priors. However, Atlas fuses input image features by direct averaging, which does not effectively model view-dependent image features or occlusion effects.

PIFu \cite{saito2019pifu} also performs multi-view fusion by averaging backprojected features, showing strong results for reconstruction of free-standing humans. However, to our knowledge, it has not been demonstrated for full, real-world scenes, which tend to introduce more complex occlusion relationships as well as semantic and geometric variety.

NeuralRecon \cite{sun2021neuralrecon} averages features only among nearby views, fusing across view clusters using a recurrent neural network (RNN). NeuralRecon achieves real-time execution, with the trade-off that incoming views must be considered sequentially. Our model lifts the constraint of sequential processing, fusing all available views jointly.

Point-MVSNet \cite{chen2019point} replaces the feature volume entirely with a feature-augmented point cloud, aggregating view features with a point cloud CNN architecture based on EdgeConv \cite{wang2019dynamic}. This is a promising approach, although point cloud learning is not as mature as regular-grid CNNs.


\noindent \textbf{Occlusion-aware MVS: }
Occlusion detection with explicit photometric and geometric constraints has traditionally played an important role in MVS \cite{kang2001handling, schonberger2016pixelwise, strecha2004wide, strecha2006combined, sun2005symmetric, yang2008stereo, zheng2014patchmatch, zitnick2000cooperative}. In addition, a number of MVS methods based on deep learning have proposed to learn visibility estimation \cite{chen2020visibility, ji2017surfacenet, ji2020surfacenet+, long2020occlusion}.

\noindent \textbf{Direct scene optimization: }
Yariv et al. \cite{yariv2020multiview} propose to directly optimize the scene representation with respect to the input images. This is effective when the target geometry is fully observed. However, it has no offline training phase in which 3D priors can be learned and then applied to new reconstructions. This prevents any significant scene completion, which is a key feature of our algorithm.

\noindent \textbf{Projective TSDF: }
In RGBD reconstruction, the projective TSDF is used as a means of approximating the true, or view-independent TSDF, by averaging together the projective TSDFs of many depth images \cite{newcombe2011kinectfusion}. It has been used as a powerful representation in its own right, as way to encode individual depth images for processing by 3D CNNs \cite{ge20173d, song2016deep}. It has also been used a reconstruction target for 3D reconstruction from single-view RGB images \cite{kim2019rgb}. In our formulation, a projective TSDF prediction acts an initial approximation of the surface geometry, which allows us to model occlusion during backprojection.


\noindent \textbf{Multi-view fusion with attention: }
For single-object reconstruction, attention has been used to fuse multiple images into a fixed-size global scene encoding \cite{xie2019pix2vox, yang2020robust, yuan2021vanet}. MVS algorithms have leveraged channel-wise attention to focus on relevant feature subspaces \cite{luo2020attention}, 2D image-space attention to aggregate visual context \cite{yang2021mvs2d, yu2021attention}, and 3D attention to promote coherence across cost volumes \cite{long2021multi}. A recent method for novel-view synthesis \cite{trevithick2020grf} has experimented with two attention mechanisms for fusing backprojected image features: AttSets \cite{yang2020robust} and Slot Attention \cite{locatello2020slot}. In our experiments, these variants do not perform as well as the transformer-based attention (see Table~\ref{table:ablations} for results and section \ref{section:ablations} for discussion).


\noindent \textbf{Transformers: }
Transformers are a family of neural network architectures that have proven very effective for sequence modeling in natural language processing \cite{devlin2018bert, vaswani2017attention}, as well as vision \cite{dosovitskiy2020image}. They are neither biased toward modeling short-range dependencies, like CNNs, nor restricted to sequential processing, like RNNs. Instead, they achieve a global receptive field by composing self-attention layers. The appeal of transformers for multi-view fusion arises from their ability to perform soft clustering of their inputs. This makes transformers a good fit for wide-baseline view fusion, which benefits from clustering views, and fusing within vs.\ across clusters in different feature subspaces.

In work submitted concurrently with ours, Aljaž et al. \cite{bozic2021transformerfusion} propose 3D reconstruction with transformers for multi-view fusion. Notably, their work further utilizes the attention weights for frame selection, to ensure that all relevant view information is considered. Our work on modeling projective occupancy is fundamentally aimed at reducing the irrelevant information, and we therefore hypothesize that these approaches may provide complimentary benefits.

\begin{figure*}
\centering
\includegraphics[width=\linewidth]{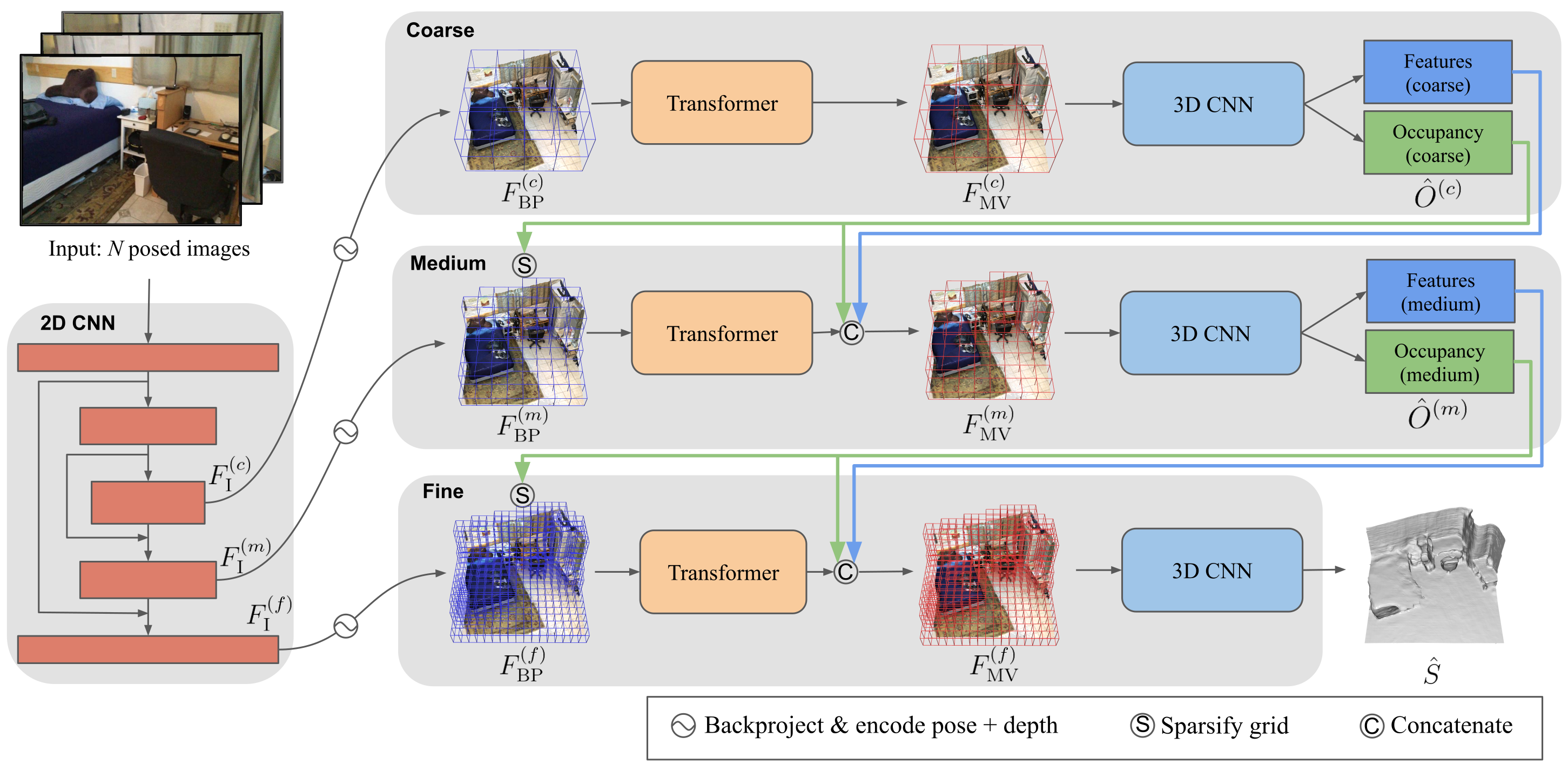}
\vspace{-0.5cm}
\caption{Model overview. A 2D CNN processes $N$ input images to produce image features at coarse, medium, and fine resolutions: $F_I^{(r)} \in \mathbb{R}^{N \times H^{(r)} \times W^{(r)} \times C^{(r)}}$, $r \in \{c, m, f\}$. At each resolution, a sparse feature volume with $V^{(r)}$ voxels is computed by backprojection, and the camera-to-voxel unit vector and depth are jointly encoded: $F_{\text{BP}}^{(r)} \in \mathbb{R}^{V^{(r)} \times N \times C^{(r)}}$. A transformer fuses image features at each voxel to produce the multi-view feature volume, $F_{\text{MV}}^{(r)} \in \mathbb{R}^{V^{(r)} \times C^{(r)}}$. At the coarse and medium resolutions, a sparse 3D CNN predicts occupancy $\hat{O}^{(r)} \in \mathbb{R}^{V^{(r)}}$ which is used to sparsify the volume. At the fine resolution, the 3D CNN predicts the final TSDF $\hat{S}$.}
\label{fig:arch}
\vspace{-0.25cm}
\end{figure*}

\section{Method}

Our goal is to predict a global TSDF volume $\hat{S}$, using an unordered sequence of input RGB images and their corresponding 6-DOF camera poses. For training, we assume the existence of ground-truth depth maps.

In broad strokes, our model extracts image features with a 2D CNN, backprojects them into a voxel grid, and predicts a TSDF with a 3D CNN. It thus bears structural similarity to existing deep volumetric reconstruction methods \cite{murez2020atlas, sun2021neuralrecon}. Sec.~\ref{section:overview} introduces the architecture overview and notation.

The first key difference from existing work is in the image feature backprojection and aggregation phase. We introduce a transformer to process single-view image features, selectively fusing them into a multi-view encoding \textit{before} aggregating per-voxel features. This significantly expands the model's ability to reason jointly about the input views, improving the localization of surfaces in its reconstructions. Details are presented in Sec.~\ref{section:transformers}.

Our second main contribution is to weight the final feature aggregation with explicitly-supervised projective occupancy predictions, enforcing that image features are only accumulated into regions near their observed surfaces. Sec.~\ref{section:projocc} expands on this component.

\subsection{Overview}
\label{section:overview}

The overall structure of our algorithm is illustrated in Fig.~\ref{fig:arch}. A 2D CNN (a feature pyramid network \cite{lin2017feature} with an MnasNet \cite{tan2019mnasnet} backbone) begins by extracting image features at coarse, medium, and fine resolutions:

\vspace{-.25cm}
\begin{equation}
    \{F_{I}^{(c)}, F_{I}^{(m)}, F_{I}^{(f)}\} = g_\theta(I), 
    \vspace{-.1cm}
\end{equation}

\noindent where $g_\theta$ is the CNN parametrized by network weights $\theta$.

At each resolution $r \in \{c, m, f\},$ the image features are backprojected onto a sparse 3D grid. This produces a feature volume, $F_{\text{BP}}^{(r)}$, in which each voxel contains a set of backprojected features, one from each image. The per-voxel features are then aggregated using our transformer and projective occupancy architecture to form a new volume, $F_{\text{MV}}^{(r)}$, containing one multi-view feature in each voxel.
A sparse 3D CNN \cite{tang2020searching} processes $F_{\text{MV}}^{(r)}$, predicting occupancy $\hat{O}^{(r)}$:

\vspace{-.15cm}
\begin{equation}
    \hat{O}^{(r)} = h_\theta^{(r)}(F_{\text{MV}}^{(r)}), \quad r \in \{c, m, f\}
    \vspace{-.15cm}
\end{equation}

\noindent where $h_\theta^{(r)}$ represents the 3D CNN at resolution $r$.

At each resolution, any voxels predicted to be unoccupied are pruned from the next, higher-resolution hierarchy level, in a coarse-to-fine manner. At the final, highest-resolution level, the TSDF $\hat{S}$ is predicted instead of occupancy, and the zero isosurface is extracted using marching cubes \cite{lorensen1987marching}. We set the voxel size at each resolution to $16\text{cm}^3$, $8\text{cm}^3$, and $4\text{cm}^3$, respectively.

In order to scale from local to full-scene reconstruction, we tile the target space with a set of non-overlapping local volumes. Then, for each tile we aim to select a diverse set of $N$ views from across the input sequence (see Sec.~\ref{section:viewselection}). Starting with the coarsest resolution, we populate $F_{\text{BP}}^{(r)}$ and $F_{\text{MV}}^{(r)}$ tile by tile. Then, we run sparse 3D convolution globally, and proceed to backprojection at the next resolution.


\begin{figure*}
\centering
\includegraphics[width=\linewidth]{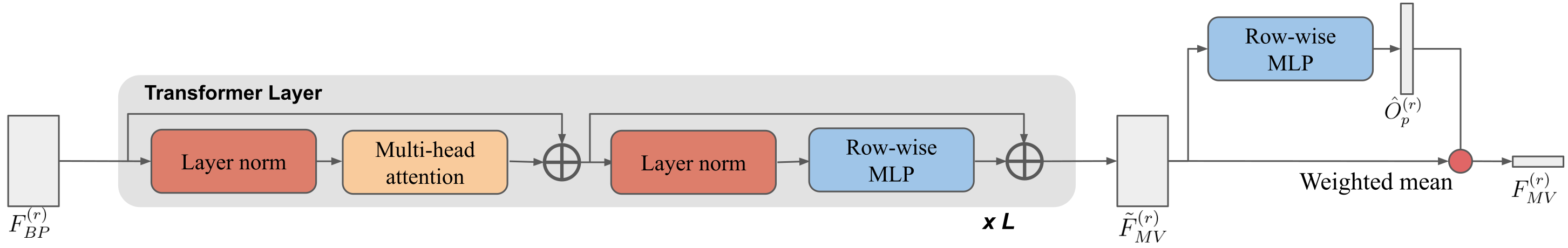}
\vspace{-0.5cm}
\caption{Our transformer architecture in detail. At each individual voxel, the transformer input features $F_{\text{BP}}^{(r)}$ from $N$ images each have channel dimension $C$. The transformer layer is repeated $L$ times, selectively fusing the inputs to produce a set of multi-view features $\tilde{F}_{MV}^{(r)}$ with the same dimensions as the input. A fully-connected layer predicts projective occupancy probabilities $\hat{O}_p^{(r)}$, which are used as weights in a final channel-wise average to produce $F_{\text{MV}}^{(r)}$.
}
\label{fig:transformer}
\vspace{-.5cm}
\end{figure*}

\subsection{Multi-view image feature fusion}
\label{section:transformers}

Our key innovation is to use a transformer to augment each backprojected single-view feature with information from other relevant views. At each voxel, the transformer takes an unordered sequence of single-view feature vectors as input, and produces a corresponding sequence of multi-view feature vectors as output:

\begin{equation}
    \tilde{F}_{\text{MV}}^{(r)} = y_\theta^{(r)}(F_{\text{BP}}^{(r)}),
\end{equation}

\noindent where $y_\theta^{(r)}$ represents the transformer at resolution $r$.

We use the tilde to indicate that $\tilde{F}_{\text{MV}}^{(r)}$ is the predecessor to $F_{\text{MV}}^{(r)}$: each voxel in $\tilde{F}_{\text{MV}}^{(r)}$ contains a sequence of multi-view features, and $F_{\text{MV}}^{(r)}$ is the result of the per-voxel feature aggregation detailed in the following section.

The correspondence between the $i^{th}$ sequence element of $F_{\text{BP}}^{(r)}$ and $\tilde{F}_{\text{MV}}^{(r)}$ is encouraged by residual connections across attention layers, and it is enforced by predicting the projective occupancy for each input view using its corresponding element of the output sequence.





Generally, the input to a transformer is an unordered sequence of feature vectors, where each feature vector is a joint encoding of the original sequence element and its position in the sequence. In our model, we replace the typical sequential positional encoding with a camera pose encoding, $\Lambda(d)$, where $d$ is the camera-to-voxel view direction unit vector and $\Lambda$ is the positional encoding from Mildenhall et al. \cite{mildenhall2020nerf}. To form the transformer input, we concatenate the image feature and the pose encoding, and reduce the resulting dimensionality with a shared fully-connected (FC) layer. We then concatenate the normalized camera-to-voxel depth and reduce with a second FC layer before applying the transformer.

Our transformer, shown in Fig. \ref{fig:transformer}, is based on the encoder part of the original transformer network introduced by Vaswani et al. \cite{vaswani2017attention}. It consists of a series of $L$ layers, where each layer contains a multi-head attention mechanism with $H$ heads, followed by a small fully-connected network.  We also employ residual connections and layer normalization within each layer. In our implementation we set $L$ and $H$ both equal to $2$. 

The following section describes the aggregation of the transformer output sequence into a single per-voxel feature vector, which is subsequently passed on to the 3D CNN.

\subsection{Projective occupancy}
\label{section:projocc}

\begin{figure}[b]
\vspace{-.5cm}
\begin{center}
\includegraphics[width=\linewidth]{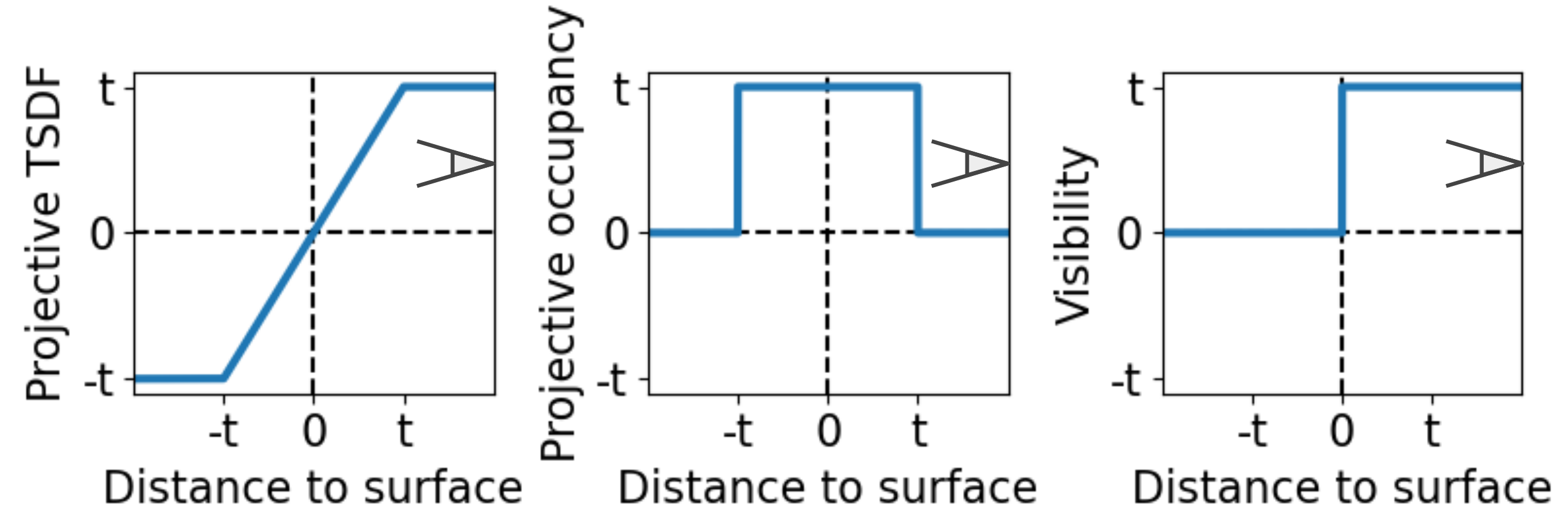}
\end{center}
\vspace{-0.2cm}
\caption{Comparison of projective distance functions, where $t$ is the TSDF truncation distance. Visibility is a function of the sign of the TSDF, and it includes observed empty space. Projective occupancy is a function of TSDF magnitude, and it describes the surface location within a margin of error. }
\label{fig:projocc}
\vspace{-.25cm}
\end{figure}

Our problem context violates key assumptions that MVS methods traditionally make, and this inspires us to re-think the notion of visibility.

Specifically, because we aim to learn view selection and fusion, we do not impose any constraints on the relative pose of the views to be fused, instead sampling broadly from across the image sequence. This results in high perspective diversity, with triangulation angles often greater than 90 degrees. This violates the typical assumptions of fronto-parallel scene structure and small baseline distance.

We therefore reconsider the notion of visibility in our context. Our goal is to place image features into 3D space such that they enable a 3D CNN to estimate the imaged surface location. If we spatially distribute those features along a camera ray according to the estimated projective occupancy, their spatial density will be centered at the estimated target surface depth. This is intuitively favorable from the perspective of the 3D CNN. In contrast, if the features are spatially distributed according to visibility, then their spatial density is spread across observed empty space, and it may not reach the true surface location if the depth is underestimated. See Fig.~\ref{fig:projocc} for an illustration. We therefore consider the projective occupancy to be a more effective prediction target for our purposes.

Furthermore, we hypothesize that it is an easier target. Fundamentally, projective occupancy requires predicting the magnitude of the TSDF, whereas visibility requires predicting the sign of the TSDF. In theory, estimating the magnitude of the TSDF at a point using two image projections can be done with only a matching cost function. However, estimating the sign of the TSDF requires understanding the direction of mismatch, and comparing it to the relative camera poses. We therefore consider the visibility to be a more difficult target, and this may contribute to the performance decrease observed in our ablation study (Table~\ref{table:ablations}, row $g$).

To introduce our projective occupancy prediction framework, we first define the projective SDF $S$,
\begin{equation}
    S = d - d_v,
\end{equation}
where $d_v$ is the camera-to-voxel depth, and $d$ is the true depth along the camera-voxel ray. We estimate $d$ in practice by projecting onto the ground truth depth map and sampling the depth at the nearest-neighbor pixel.

The projective occupancy $O_p$ can then be obtained by thresholding the absolute value of $S$ on the truncation distance $t$:
\begin{equation}
    O_p = \left\{
        \begin{array}{ll}
            1 & \quad |S| < t \\
            0 & \quad |S| \geq t
        \end{array}
    \right..
\end{equation}

Our model estimates the projective occupancy likelihood $X \in \mathbb{R}^N$ as
\begin{equation}
    X = z_\theta^{(r)}(\tilde{F}_{\text{MV}}^{(r)}),
\end{equation}
\noindent where $z_\theta^{(r)}$ is a single, shared, FC layer at resolution $r$. In order to supervise $X$, a sigmoid is applied to produce the projective occupancy probabilities:
\begin{equation}
    \hat{O}_p^{(r)} = \sigma(X)   
\end{equation}
Then a loss is computed as binary cross-entropy between $\hat{O}_p^{(r)}$ and the groundtruth projective occupancy.

In order to use $X$ to inform feature aggregation, we concatenate a zero-likelihood to $X$ and apply a softmax to compute a weight vector $W \in \mathbb{R}^{1 \times N+1}$. We then concatenate a zero feature vector to $\tilde{F}_{\text{MV}}^{(r)}$, resulting in dimensions $(N+1) \times C$, and reduce with a weighted sum:
\vspace{-.15cm}
\begin{equation}
    F_{\text{MV}}^{(r)} = W\tilde{F}_{\text{MV}}^{(r)}
\vspace{-.15cm}
\end{equation}
The softmax weight normalization ensures that the distribution of $F_{\text{MV}}^{(r)}$ is invariant to the number of input views. The zero-padding of both features and likelihoods causes $F_{\text{MV}}^{(r)}$ to be near zero if all the predicted occupancy likelihoods are low.

\subsection{View selection}
\label{section:viewselection}

Our method does not depend on heuristics to select optimally positioned input views. Conversely, we aim to train our model on an unconstrained set of views that is as diverse as possible while remaining computationally tractable, such that it can learn to fuse features across the appropriate range of pose differences. We employ heuristics only to reduce the overall number of views while maintaining diversity.

To this end, we first remove redundant views by applying the keyframe selection strategy from Sun et al. \cite{sun2021neuralrecon}. Then, for each local sub-volume, we select $N$ views via uniform random sampling from among the remaining views whose camera frustums intersect the target volume. During training we set $N = 20$, and during testing we set $N = 60$. For redundant frame removal, we set $R_{max}$ to $15$ degrees, and we set $t_{max}$ to 0.1 m for training and 0.2 m for testing.




\begin{figure*}[th]
\begin{center}
\includegraphics[width=\linewidth]{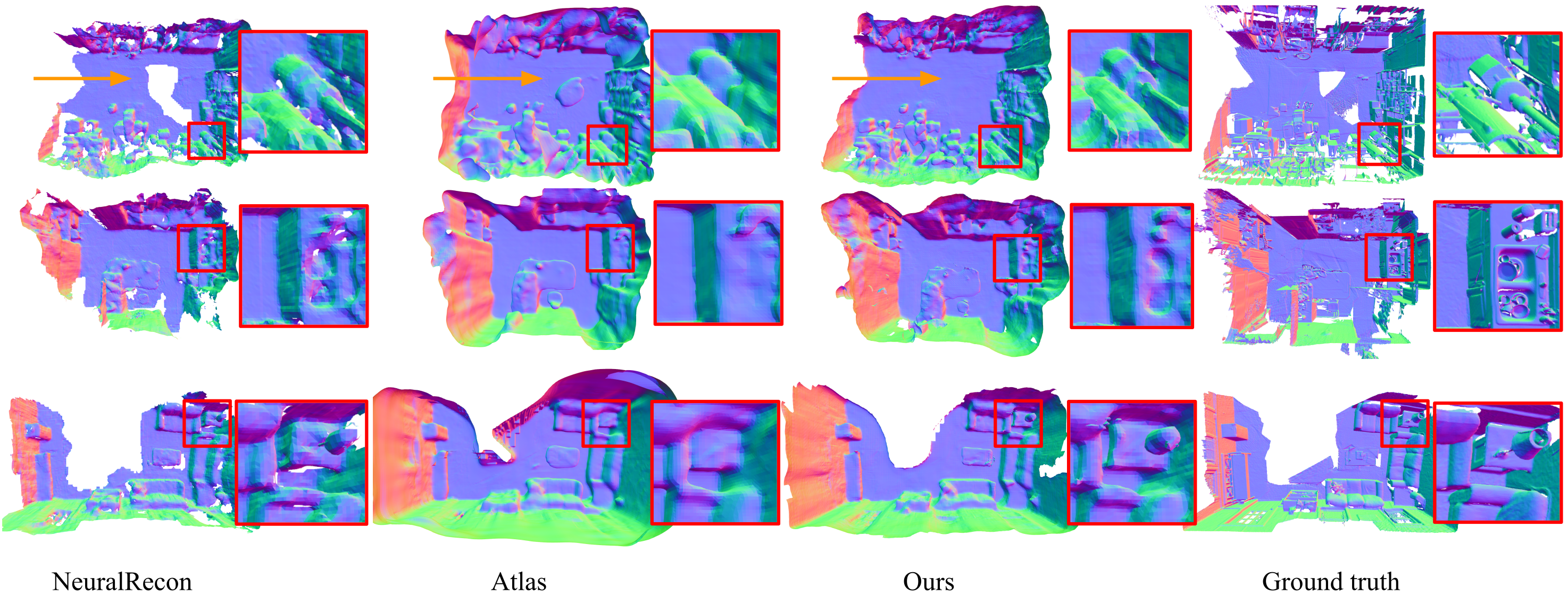}
\end{center}
\vspace{-0.3cm}
\caption{Qualitative results on ScanNet. The inset boxes show enlarged regions where our model reconstructs a high degree of detail. With the orange arrows, we highlight another strength of our model: it fills in unobserved regions plausibly, without leaving holes or artifacts.}
\label{fig:qualitative}
\vspace{-.5cm}
\end{figure*}

\subsection{Training}

\noindent \textbf{Loss function: }The projective occupancy loss $\lambda_P^{(r)}$ at each hierarchy level, and the occupancy loss at the coarser levels $\lambda_O^{(r)}$, are computed using binary cross-entropy. The TSDF loss at the finest level, $\lambda_S$, is computed by $l1$ distance to the ground truth, after log-transforming the prediction and ground truth following \cite{dai2017shape}. Then the total loss $L$ is
\vspace{-.1cm}
\begin{align*}
    L = \lambda_P^{(c)} + \lambda_P^{(m)} + \lambda_P^{(f)} + \lambda_O^{(c)} + \lambda_O^{(m)} + \lambda_S
\end{align*}
\noindent \textbf{Ground truth: }We compute our fine-resolution reconstruction target using TSDF fusion at 4 cm resolution, discarding all measurements greater than 3 m due to sensor noise at longer ranges. We then threshold that TSDF on the truncation distance to obtain a fine-resolution occupancy volume, which we downsample by morphological dilation to produce the medium and coarse reconstruction targets. As in Murez et al. \cite{murez2020atlas}, we mark any column of the ground truth TSDF volume as unoccupied if it is entirely unobserved.

During training we select sub-volumes by randomly selecting TSDF subcrops with size $96 \times 96 \times 48$ voxels, or $3.84~\text{m} \times 3.84~\text{m} \times 1.92~\text{m}$. We augment with random horizontal reflections and rotations about the gravitational axis.

\noindent \textbf{Training phases: }During our initial training phase, the projective occupancy predictions are supervised, but they are not otherwise used: the transformer output sequence is aggregated with an unweighted average. This aids stability. Also during this phase, the 2D CNN weights, which are pre-trained on ImageNet, are frozen. The learning rate is $10^{-3}$, the batch size is $4$, and this phase lasts 300 epochs.

In the second phase, the projective occupancy predictions are used for weighted-average aggregation of the transformer outputs, as shown in Fig.~\ref{fig:transformer}. In addition, the 2D CNN weights are unfrozen, except for the batch norm weights and statistics. The learning rate is lowered to $10^{-4}$, the batch size is lowered to $2$. This phase lasts 100 epochs.

\noindent \textbf{Implementation details: }
We use the Adam optimizer with $\beta_1 = 0.9$, $\beta_2 = 0.999$, $\epsilon = 10^{-8}$, and a linear learning rate warm-up from $0$ over $2,000$ steps. Training takes approximately 84 hours on a single Nvidia RTX 3090 graphics card. We implement our model in PyTorch, using the PyTorch Lightning framework \cite{falcon2019pl}. We use torchsparse \cite{tang2020searching} for our sparse 3D CNN, and Open3D \cite{Zhou2018o3d} for visualization and geometry processing. During training, we randomly drop out voxels to reduce memory cost, following \cite{sun2021neuralrecon}.

\section{Experiments}
\label{section:evaluation}

For all experiments, we train our method on the ScanNet dataset \cite{dai2017scannet}: 1,513 RGBD scans of 707 indoor spaces. We use the official train/validation/test split.

For quantitative comparison, we compute a set of 3D metrics as defined by Murez et al. \cite{murez2020atlas}. To avoid penalizing the volumetric methods for filling in areas that are not present in the ground truth, we trim the reconstructed mesh to within the observed regions. To do this, we render the ground-truth mesh to a set of depth maps $D$ from the perspective of each camera pose. Then we render the predicted mesh to a set of depth maps $\hat{D}$. We mask out pixels in $\hat{D}$ that do not have a valid depth in $D$, and re-fuse the masked predicted depth into a trimmed mesh via TSDF fusion.


\begin{table}[b]
\vspace{-.5cm}
\small
\begin{center}
\begin{tabular}{|c c c c c c|}
\thickhline
& Acc $\downarrow$ & Comp $\downarrow$ & Prec $\uparrow$ & Recall $\uparrow$ & F-score $\uparrow$ \\
\thickhline
\sc{\textbf{ScanNet}} & & & & & \\
Atlas & 0.068 & 0.098 & 0.640 & 0.539 & 0.583 \\
NeuralRecon & \textbf{0.049} & 0.133 & 0.691 & 0.461 & 0.551 \\
Ours & 0.054 & \textbf{0.090} & \textbf{0.708} & \textbf{0.588} & \textbf{0.641} \\
\hline
\sc{\textbf{ICL-NUIM}} & & & & & \\
Atlas & 0.175 & 0.314 & 0.280 & 0.194 & 0.229 \\
NeuralRecon & 0.215 & 1.031 & 0.214 & 0.036 & 0.058  \\
Ours & \textbf{0.102} & \textbf{0.146} & \textbf{0.449} & \textbf{0.375} & \textbf{0.408} \\
\hline
\sc{\textbf{TUM-RGBD}} & & & & & \\
Atlas & 0.208 & 2.344 & 0.360 & 0.089 & 0.132 \\
NeuralRecon & \textbf{0.130} & 2.528 & \textbf{0.382} & 0.075 & 0.115 \\
Ours & 0.175 & \textbf{0.314} & 0.280 & \textbf{0.194} & \textbf{0.229} \\
\hline
\end{tabular}
\end{center}
\vspace{-0.2cm}
\caption{Reconstruction metrics (as defined as in \cite{murez2020atlas}), comparison with volumetric methods.}
\label{table:scannet_results_vol}
\end{table}

\subsection{Volumetric baselines}

Our primary comparison is with algorithms that, like ours, can complete geometry in unobserved regions. These are the deep volumetric methods, Atlas \cite{murez2020atlas} and NeuralRecon \cite{sun2021neuralrecon}, and we use the provided pre-trained models. For Atlas, we select every $5^{\text{th}}$ frame as input, and for NeuralRecon we use the frame selection proposed by its authors. We evaluate on the ScanNet test set (100 scenes), the ICL-NUIM dataset (8 scenes), and the TUM-RGBD dataset (13 scenes). For ScanNet, we evaluate against the provided ground-truth meshes. For TUM-RGBD and ICL-NUIM, we generate ground truth by TSDF fusion at 4 cm resolution.

Quantitative results are shown in Table \ref{table:scannet_results_vol}. We consider F-score to be the most important metric, as it captures the trade-off between precision and recall. Our F-score indicates a significant improvement over state-of-the-art methods. We also report the accuracy of our projective occupancy predictions at each resolution in Table \ref{table:projocc}, and we compare against the default prediction of $true$ everywhere.

Qualitative results are shown in Fig.~\ref{fig:qualitative}. We observe increased accuracy relative to the baseline methods, particularly in areas with many small objects and a high degree of occlusion, such as cluttered countertops. In these regions, our model produces a high level of detail while also filling in holes arising from occlusion (Fig.~\ref{fig:qualitative}, rows 1 and 2). We note that in large unobserved regions (Fig.~\ref{fig:qualitative}, row 3), our model's performance degrades gracefully: whereas Atlas tends to incorrectly place walls at the boundary, and NeuralRecon typically does not produce any geometry, VoRTX extends observed surfaces for a plausible distance without introducing large artifacts. 

\begin{figure}[t]
\begin{center}
\includegraphics[width=\linewidth]{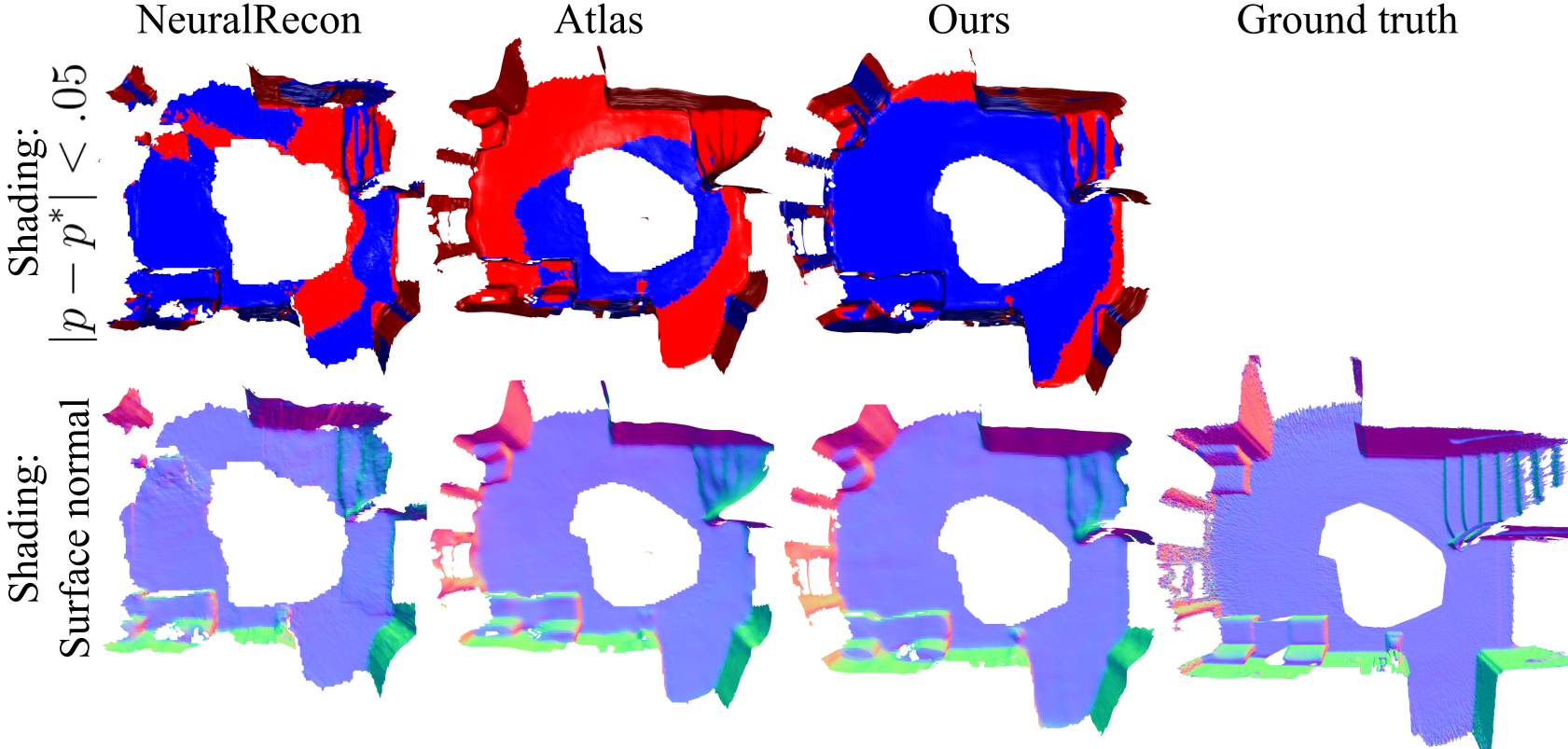}
\end{center}
\vspace{-0.4cm}
\caption{Trimmed mesh predictions (see Sec.~\ref{section:evaluation}). Top: shaded blue for predicted vertices $p^*$ within $5cm$ of a ground-truth vertex $p$, red otherwise. Bottom: shaded by surface normal. Our results show improved accuracy, even in cases with similar visual quality.}
\label{fig:meshcoloring}
\vspace{-.5cm}
\end{figure}

We also observe that in many cases, even when reconstruction quality is visually similar, our model localizes surfaces more accurately, as shown in Fig.~\ref{fig:meshcoloring}.

\begin{table}[b]
\vspace{-.5cm}
\small
\begin{center}
\begin{tabular}{|c|c|c c c|}
\thickhline
Hierarchy Lvl. & Proj. Occ. Prediction & Prec $\uparrow$ & Recall $\uparrow$ & Acc $\uparrow$ \\
\thickhline
\multirow{2}{*}{4cm} & Default (true everywhere) & 0.237 & 1.000 & 0.237 \\
& Ours & 0.702 & 0.347 & 0.813 \\
\hline
\multirow{2}{*}{8cm} & Default (true everywhere) & 0.301 & 1.000 &  0.301 \\
& Ours & 0.750 & 0.627 & 0.829 \\
\hline
\multirow{2}{*}{16cm} & Default (true everywhere) & 0.067 & 1.000 & 0.067 \\
& Ours & 0.739 & 0.661 & 0.961 \\
\hline
\end{tabular}
\end{center}
\vspace{-0.2cm}
\caption{Projective occupancy results. The default behavior is to assume projective occupancy is \textit{true} for all voxels.}
\label{table:projocc}
\end{table}

\subsection{Depth-prediction baselines}

For completeness, we compare with deep MVS networks that estimate depth maps, reconstructing only observed surfaces: DeepVideoMVS (with fusion) \cite{duzceker2021deepvideomvs}, Fast-MVSNet \cite{yu2020fmvs}, GPMVS (batched) \cite{hou2019multi}, and Point-MVSNet \cite{chen2019point}. For DeepVideoMVS, we use the ScanNet pre-trained weights. For Fast-MVSNet, GPMVS, and Point-MVSNet, we fine-tune on ScanNet, starting from the pre-trained models. For Point-MVSNet and Fast-MVSNet, we modify the parameters for the longer ranges in ScanNet relative to DTU \cite{aanaes2016large}: we use 96 depth hypotheses, every 5 cm starting at 50 cm. We fuse predicted depths into point clouds following \cite{galliani2015massively}. For all depth-prediction methods, we select views following Duzceker et al. \cite{duzceker2021deepvideomvs}, using four source images for each reference image. As shown in Table~\ref{table:scannet_results_depth}, VoRTX produces higher F-scores, indicating that it does not compromise on observed surfaces in order to complete unobserved regions.

\subsection{Ablation experiments}
\label{section:ablations}

\begin{table}[b]
\vspace{-.25cm}
\small
\begin{center}
\begin{tabular}{|c c c c c c|}
\thickhline
& Acc $\downarrow$ & Comp $\downarrow$ & Prec $\uparrow$ & Recall $\uparrow$ & F-score $\uparrow$ \\
\thickhline
DeepVideoMVS & 0.079 & 0.133 & 0.521 & 0.454 & 0.474 \\
Fast-MVSNet  & 0.042 & 0.225 & 0.746 & 0.383 & 0.495 \\
GPMVS        & 0.066 & 0.117 & 0.591 & 0.513 & 0.539 \\
Point-MVSNet & \textbf{0.037} & 0.278 & \textbf{0.790} & 0.363 & 0.484 \\
Ours         & 0.054 & \textbf{0.090} & 0.708 & \textbf{0.588} & \textbf{0.641} \\
\hline
\end{tabular}
\end{center}
\vspace{-0.2cm}
\caption{ScanNet reconstruction metrics (as defined as in \cite{murez2020atlas}), comparison with depth-prediction methods.}
\label{table:scannet_results_depth}
\end{table}

\begin{table}[b]
\vspace{-.25cm}
\footnotesize
\begin{center}
\begin{tabular}{|c|c c c|c c c c c|}
\thickhline
& Transf. & Proj.Occ. & Pose & Acc $\downarrow$ & Comp $\downarrow$ & Prec $\uparrow$ & Rec $\uparrow$ & F-score $\uparrow$ \\
\thickhline
a & \checkmark & \checkmark & \checkmark & 0.054 & 0.090 & 0.708 & 0.588 & 0.641 \\
\hline
b & \checkmark & & \checkmark & 0.058 & 0.090 & 0.681 & 0.579 & 0.624 \\
\hline
c & & \checkmark & \checkmark & 0.067 & 0.110 & 0.626 & 0.510 & 0.560 \\
\hline
d & & & \checkmark & 0.071 & 0.125 & 0.611 & 0.487 & 0.540 \\
\hline
e & \checkmark & \checkmark & & 0.053 & 0.091 & 0.701 & 0.579 & 0.633 \\
\hline
f & L=1, H=1 & \checkmark & \checkmark & 0.057 & 0.090 & 0.684 & 0.572 & 0.622 \\
\hline
g & \checkmark & Vis. & \checkmark & 0.057 & 0.089 & 0.677 & 0.562 & 0.613 \\
\hline
h & AttSets & \checkmark & \checkmark & 0.057 & 0.098 & 0.680 & 0.563 & 0.614 \\
\hline
i & Slot Attn. & \checkmark & \checkmark & 0.075 & 0.210 & 0.546 & 0.346 & 0.420 \\
\hline
\end{tabular}
\end{center}
\vspace{-0.2cm}
\caption{Ablation experiments on ScanNet.}
\label{table:ablations}
\end{table}

In Table \ref{table:ablations} we present ablation experiments to validate our model. In each, the model architecture is modified and re-trained from scratch. Row \textit{a} is VoRTX, unmodified.

\noindent \textbf{Transformer: }
We first experiment with removing the transformer entirely (row \textit{c}). In this case, projective occupancy predictions are made on the basis of the single-view features, aggregating by weighted average. This causes a significant drop in F-score. We also experiment with removing both transformer and projective occupancy (\textit{d}), aggregating within voxels by unweighted average. This causes a further F-score drop. We conclude that the transformer is responsible for most of VoRTX's performance gain.

In \textit{f} we alter the hyperparameters of the transformer, using only a single layer and a single attention head, resulting in a moderate F-score decrease. We thus hypothesize that additional layers may lead to further performance gains.

In \textit{h} and \textit{i}, we replace the transformer with alternative attention mechanisms, following GRF \cite{trevithick2020grf}. The projective occupancy is predicted using single-view features. In \textit{h}, the AttSets \cite{yang2020robust} model shows a moderate F-score decrease. This may be due to the fact that AttSets has only one attention layer, or that it doesn't model pairwise attention between views. In \textit{i}, using Slot Attention \cite{locatello2020slot}, our model does not converge well during training, and further investigation may be required to fully characterize the technique.

\noindent \textbf{Projective occupancy: }
We also experiment with removing the projective occupancy prediction while keeping the transformer, aggregating the transformer outputs by direct averaging (\textit{b}). In \textit{g}, we keep the same architecture, but we supervise $\hat{O}_{p}^{(r)}$ with the visibility instead of projective occupancy. In both cases we see a small performance decrease, supporting our hypotheses that the model benefits from supervising the aggregation weights, and that projective occupancy is a more effective weighting function than visibility.

\noindent \textbf{Pose: }
In \textit{e}, the model does not encode pose information into the image features during backprojection (it does still encode camera-to-voxel depth). This results in only a very slight performance decrease. We interpret this to suggest that although the viewing direction is useful information, most of its benefit can be obtained with attention-based comparison of pose-agnostic image features.

\subsection{Inference time}

Our method achieves speeds compatible with interactive applications on commodity hardware. We benchmark VoRTX on the ScanNet test set, using an AMD Threadripper 2950X and an NVIDIA RTX 3090. It averages 14.2 FPS, counting only selected keyframes.

\section{Limitations}

Because VoRTX uses a voxel representation, it is subject to a trade-off between resolution and memory use. We use 4 cm voxels, which are acceptable for indoor scenes but can cause aliasing for thin structures. In addition, reflective surfaces are often missing from our reconstructions. We believe this is partially due to the failure of the depth sensors for those surfaces, leading to gaps in supervision.

\section{Conclusion}

We have presented a novel method for multi-view fusion using transformers, applied toward deep volumetric MVS. We show that this produces better reconstructions than state-of-the-art methods on ScanNet, TUM-RGBD, and ICL-NUIM. Our model is trained only on ScanNet, generalizing well to the two other datasets without fine-tuning. Our projective occupancy framework opens the door to occlusion-awareness for deep volumetric MVS.

In the future, a focus on thin structures and reflective surfaces could yield improvements. Use of simulated training data, or alternative depth sensors, may facilitate learning and open possibilities for new data domains. Further attention to scalability may be beneficial for transferring to large-scale reconstructions. Finally, we anticipate that the transformer-based view fusion may also be applicable to tasks such as fusing multiple sensing modalities. \\





\section{Acknowledgements} Support for this work was provided by ONR grants N00014-19-1-2553 and N00174-19-1-0024, as well as NSF grants 1911230 and OAC-1925717.

{\small
\bibliographystyle{ieee_fullname}
\bibliography{references}

\begin{thebibliography}{10}\itemsep=-1pt

\bibitem{aanaes2016large}
Henrik Aan{\ae}s, Rasmus~Ramsb{\o}l Jensen, George Vogiatzis, Engin Tola, and
  Anders~Bjorholm Dahl.
\newblock Large-scale data for multiple-view stereopsis.
\newblock {\em International Journal of Computer Vision}, 120(2):153--168,
  2016.

\bibitem{bozic2021transformerfusion}
Alja{\v{z}} Bo{\v{z}}i{\v{c}}, Pablo Palafox, Justus Thies, Angela Dai, and
  Matthias Nie{\ss}ner.
\newblock {TransformerFusion}: Monocular {RGB} scene reconstruction using
  transformers.
\newblock {\em Proc. Neural Information Processing Systems (NeurIPS)}, 2021.

\bibitem{chen2020visibility}
Rui Chen, Songfang Han, Jing Xu, et~al.
\newblock Visibility-aware point-based multi-view stereo network.
\newblock {\em IEEE transactions on pattern analysis and machine intelligence},
  2020.

\bibitem{chen2019point}
Rui Chen, Songfang Han, Jing Xu, and Hao Su.
\newblock Point-based multi-view stereo network.
\newblock In {\em Proceedings of the IEEE/CVF International Conference on
  Computer Vision}, pages 1538--1547, 2019.

\bibitem{dai2017scannet}
Angela Dai, Angel~X Chang, Manolis Savva, Maciej Halber, Thomas Funkhouser, and
  Matthias Nie{\ss}ner.
\newblock {ScanNet}: Richly-annotated {3D} reconstructions of indoor scenes.
\newblock In {\em Proceedings of the IEEE conference on computer vision and
  pattern recognition}, pages 5828--5839, 2017.

\bibitem{dai2018scancomplete}
Angela Dai, Daniel Ritchie, Martin Bokeloh, Scott Reed, J{\"u}rgen Sturm, and
  Matthias Nie{\ss}ner.
\newblock {ScanComplete}: Large-scale scene completion and semantic
  segmentation for {3D} scans.
\newblock In {\em Proceedings of the IEEE Conference on Computer Vision and
  Pattern Recognition}, pages 4578--4587, 2018.

\bibitem{dai2017shape}
Angela Dai, Charles Ruizhongtai~Qi, and Matthias Nie{\ss}ner.
\newblock Shape completion using 3d-encoder-predictor cnns and shape synthesis.
\newblock In {\em Proceedings of the IEEE Conference on Computer Vision and
  Pattern Recognition}, pages 5868--5877, 2017.

\bibitem{devlin2018bert}
Jacob Devlin, Ming-Wei Chang, Kenton Lee, and Kristina Toutanova.
\newblock Bert: Pre-training of deep bidirectional transformers for language
  understanding.
\newblock {\em arXiv preprint arXiv:1810.04805}, 2018.

\bibitem{dosovitskiy2020image}
Alexey Dosovitskiy, Lucas Beyer, Alexander Kolesnikov, Dirk Weissenborn,
  Xiaohua Zhai, Thomas Unterthiner, Mostafa Dehghani, Matthias Minderer, Georg
  Heigold, Sylvain Gelly, et~al.
\newblock An image is worth 16x16 words: Transformers for image recognition at
  scale.
\newblock {\em arXiv preprint arXiv:2010.11929}, 2020.

\bibitem{duzceker2021deepvideomvs}
Arda Duzceker, Silvano Galliani, Christoph Vogel, Pablo Speciale, Mihai
  Dusmanu, and Marc Pollefeys.
\newblock {DeepVideoMVS}: Multi-view stereo on video with recurrent
  spatio-temporal fusion.
\newblock In {\em Proceedings of the IEEE/CVF Conference on Computer Vision and
  Pattern Recognition}, pages 15324--15333, 2021.

\bibitem{falcon2019pl}
WA Falcon and et al.
\newblock {PyTorch} lightning.
\newblock {\em GitHub. Note:
  https://github.com/PyTorchLightning/pytorch-lightning}, 3, 2019.

\bibitem{galliani2015massively}
Silvano Galliani, Katrin Lasinger, and Konrad Schindler.
\newblock Massively parallel multiview stereopsis by surface normal diffusion.
\newblock In {\em Proceedings of the IEEE International Conference on Computer
  Vision}, pages 873--881, 2015.

\bibitem{ge20173d}
Liuhao Ge, Hui Liang, Junsong Yuan, and Daniel Thalmann.
\newblock {3D} convolutional neural networks for efficient and robust hand pose
  estimation from single depth images.
\newblock In {\em Proceedings of the IEEE Conference on Computer Vision and
  Pattern Recognition}, pages 1991--2000, 2017.

\bibitem{hou2019multi}
Yuxin Hou, Juho Kannala, and Arno Solin.
\newblock Multi-view stereo by temporal nonparametric fusion.
\newblock In {\em Proceedings of the IEEE/CVF International Conference on
  Computer Vision}, pages 2651--2660, 2019.

\bibitem{huang2018deepmvs}
Po-Han Huang, Kevin Matzen, Johannes Kopf, Narendra Ahuja, and Jia-Bin Huang.
\newblock {DeepMVS}: Learning multi-view stereopsis.
\newblock In {\em Proceedings of the IEEE Conference on Computer Vision and
  Pattern Recognition}, pages 2821--2830, 2018.

\bibitem{im2019dpsnet}
Sunghoon Im, Hae-Gon Jeon, Stephen Lin, and In~So Kweon.
\newblock {DPSNet}: End-to-end deep plane sweep stereo.
\newblock {\em arXiv preprint arXiv:1905.00538}, 2019.

\bibitem{ji2017surfacenet}
Mengqi Ji, Juergen Gall, Haitian Zheng, Yebin Liu, and Lu Fang.
\newblock {SurfaceNet}: An end-to-end {3D} neural network for multiview
  stereopsis.
\newblock In {\em Proceedings of the IEEE International Conference on Computer
  Vision}, pages 2307--2315, 2017.

\bibitem{ji2020surfacenet+}
Mengqi Ji, Jinzhi Zhang, Qionghai Dai, and Lu Fang.
\newblock {SurfaceNet+}: An end-to-end {3D} neural network for very sparse
  multi-view stereopsis.
\newblock {\em IEEE Transactions on Pattern Analysis and Machine Intelligence},
  2020.

\bibitem{kang2001handling}
Sing~Bing Kang, Richard Szeliski, and Jinxiang Chai.
\newblock Handling occlusions in dense multi-view stereo.
\newblock In {\em Proceedings of the 2001 IEEE Computer Society Conference on
  Computer Vision and Pattern Recognition. CVPR 2001}, volume~1, pages I--I.
  IEEE, 2001.

\bibitem{kim2019rgb}
Hanjun Kim, Jiyoun Moon, and Beomhee Lee.
\newblock {RGB-to-TSDF}: Direct {TSDF} prediction from a single {RGB} image for
  dense {3D} reconstruction.
\newblock In {\em 2019 IEEE/RSJ International Conference on Intelligent Robots
  and Systems (IROS)}, pages 6714--6720. IEEE, 2019.

\bibitem{lin2017feature}
Tsung-Yi Lin, Piotr Doll{\'a}r, Ross Girshick, Kaiming He, Bharath Hariharan,
  and Serge Belongie.
\newblock Feature pyramid networks for object detection.
\newblock In {\em Proceedings of the IEEE conference on computer vision and
  pattern recognition}, pages 2117--2125, 2017.

\bibitem{locatello2020slot}
Francesco Locatello, Dirk Weissenborn, Thomas Unterthiner, Aravindh Mahendran,
  Georg Heigold, Jakob Uszkoreit, Alexey Dosovitskiy, and Thomas Kipf.
\newblock Object-centric learning with slot attention.
\newblock In H. Larochelle, M. Ranzato, R. Hadsell, M.~F. Balcan, and H. Lin,
  editors, {\em Advances in Neural Information Processing Systems}, volume~33,
  pages 11525--11538. Curran Associates, Inc., 2020.

\bibitem{long2021multi}
Xiaoxiao Long, Lingjie Liu, Wei Li, Christian Theobalt, and Wenping Wang.
\newblock Multi-view depth estimation using epipolar spatio-temporal networks.
\newblock In {\em Proceedings of the IEEE/CVF Conference on Computer Vision and
  Pattern Recognition}, pages 8258--8267, 2021.

\bibitem{long2020occlusion}
Xiaoxiao Long, Lingjie Liu, Christian Theobalt, and Wenping Wang.
\newblock Occlusion-aware depth estimation with adaptive normal constraints.
\newblock In {\em European Conference on Computer Vision}, pages 640--657.
  Springer, 2020.

\bibitem{lorensen1987marching}
William~E Lorensen and Harvey~E Cline.
\newblock Marching cubes: A high resolution 3d surface construction algorithm.
\newblock {\em ACM siggraph computer graphics}, 21(4):163--169, 1987.

\bibitem{luo2020attention}
Keyang Luo, Tao Guan, Lili Ju, Yuesong Wang, Zhuo Chen, and Yawei Luo.
\newblock Attention-aware multi-view stereo.
\newblock In {\em Proceedings of the IEEE/CVF Conference on Computer Vision and
  Pattern Recognition}, pages 1590--1599, 2020.

\bibitem{mildenhall2020nerf}
Ben Mildenhall, Pratul~P Srinivasan, Matthew Tancik, Jonathan~T Barron, Ravi
  Ramamoorthi, and Ren Ng.
\newblock {NeRF}: Representing scenes as neural radiance fields for view
  synthesis.
\newblock In {\em European conference on computer vision}, pages 405--421.
  Springer, 2020.

\bibitem{murez2020atlas}
Zak Murez, Tarrence van As, James Bartolozzi, Ayan Sinha, Vijay Badrinarayanan,
  and Andrew Rabinovich.
\newblock Atlas: End-to-end {3D} scene reconstruction from posed images.
\newblock In {\em Computer Vision--ECCV 2020: 16th European Conference,
  Glasgow, UK, August 23--28, 2020, Proceedings, Part VII 16}, pages 414--431.
  Springer, 2020.

\bibitem{newcombe2011kinectfusion}
Richard~A Newcombe, Shahram Izadi, Otmar Hilliges, David Molyneaux, David Kim,
  Andrew~J Davison, Pushmeet Kohi, Jamie Shotton, Steve Hodges, and Andrew
  Fitzgibbon.
\newblock {KinectFusion}: Real-time dense surface mapping and tracking.
\newblock In {\em 2011 10th IEEE international symposium on mixed and augmented
  reality}, pages 127--136. IEEE, 2011.

\bibitem{ronneberger2015u}
Olaf Ronneberger, Philipp Fischer, and Thomas Brox.
\newblock {U-Net}: Convolutional networks for biomedical image segmentation.
\newblock In {\em International Conference on Medical image computing and
  computer-assisted intervention}, pages 234--241. Springer, 2015.

\bibitem{saito2019pifu}
Shunsuke Saito, Zeng Huang, Ryota Natsume, Shigeo Morishima, Angjoo Kanazawa,
  and Hao Li.
\newblock {PIFu}: Pixel-aligned implicit function for high-resolution clothed
  human digitization.
\newblock In {\em Proceedings of the IEEE/CVF International Conference on
  Computer Vision}, pages 2304--2314, 2019.

\bibitem{schonberger2016pixelwise}
Johannes~L Sch{\"o}nberger, Enliang Zheng, Jan-Michael Frahm, and Marc
  Pollefeys.
\newblock Pixelwise view selection for unstructured multi-view stereo.
\newblock In {\em European Conference on Computer Vision}, pages 501--518.
  Springer, 2016.

\bibitem{song2016deep}
Shuran Song and Jianxiong Xiao.
\newblock Deep sliding shapes for amodal {3D} object detection in {RGB-D}
  images.
\newblock In {\em Proceedings of the IEEE conference on computer vision and
  pattern recognition}, pages 808--816, 2016.

\bibitem{strecha2004wide}
Christoph Strecha, Rik Fransens, and Luc Van~Gool.
\newblock Wide-baseline stereo from multiple views: a probabilistic account.
\newblock In {\em Proceedings of the 2004 IEEE Computer Society Conference on
  Computer Vision and Pattern Recognition, 2004. CVPR 2004.}, volume~1, pages
  I--I. IEEE, 2004.

\bibitem{strecha2006combined}
Christoph Strecha, Rik Fransens, and Luc Van~Gool.
\newblock Combined depth and outlier estimation in multi-view stereo.
\newblock In {\em 2006 IEEE Computer Society Conference on Computer Vision and
  Pattern Recognition (CVPR'06)}, volume~2, pages 2394--2401. IEEE, 2006.

\bibitem{sun2005symmetric}
Jian Sun, Yin Li, Sing~Bing Kang, and Heung-Yeung Shum.
\newblock Symmetric stereo matching for occlusion handling.
\newblock In {\em 2005 IEEE Computer Society Conference on Computer Vision and
  Pattern Recognition (CVPR'05)}, volume~2, pages 399--406. IEEE, 2005.

\bibitem{sun2021neuralrecon}
Jiaming Sun, Yiming Xie, Linghao Chen, Xiaowei Zhou, and Hujun Bao.
\newblock {NeuralRecon}: Real-time coherent {3D} reconstruction from monocular
  video.
\newblock In {\em Proceedings of the IEEE/CVF Conference on Computer Vision and
  Pattern Recognition}, pages 15598--15607, 2021.

\bibitem{tan2019mnasnet}
Mingxing Tan, Bo Chen, Ruoming Pang, Vijay Vasudevan, Mark Sandler, Andrew
  Howard, and Quoc~V Le.
\newblock {MnasNet}: Platform-aware neural architecture search for mobile.
\newblock In {\em Proceedings of the IEEE/CVF Conference on Computer Vision and
  Pattern Recognition}, pages 2820--2828, 2019.

\bibitem{tang2020searching}
Haotian Tang, Zhijian Liu, Shengyu Zhao, Yujun Lin, Ji Lin, Hanrui Wang, and
  Song Han.
\newblock Searching efficient {3D} architectures with sparse point-voxel
  convolution.
\newblock In {\em European Conference on Computer Vision}, pages 685--702.
  Springer, 2020.

\bibitem{tola2009daisy}
Engin Tola, Vincent Lepetit, and Pascal Fua.
\newblock Daisy: An efficient dense descriptor applied to wide-baseline stereo.
\newblock {\em IEEE transactions on pattern analysis and machine intelligence},
  32(5):815--830, 2009.

\bibitem{trevithick2020grf}
Alex Trevithick and Bo Yang.
\newblock Grf: Learning a general radiance field for 3d scene representation
  and rendering.
\newblock {\em arXiv preprint arXiv:2010.04595}, 2020.

\bibitem{vaswani2017attention}
Ashish Vaswani, Noam Shazeer, Niki Parmar, Jakob Uszkoreit, Llion Jones,
  Aidan~N Gomez, {\L}ukasz Kaiser, and Illia Polosukhin.
\newblock Attention is all you need.
\newblock In {\em Advances in neural information processing systems}, pages
  5998--6008, 2017.

\bibitem{wang2018mvdepthnet}
Kaixuan Wang and Shaojie Shen.
\newblock {MVDepthNet}: Real-time multiview depth estimation neural network.
\newblock In {\em 2018 International conference on {3D} vision (3DV)}, pages
  248--257. IEEE, 2018.

\bibitem{wang2019dynamic}
Yue Wang, Yongbin Sun, Ziwei Liu, Sanjay~E Sarma, Michael~M Bronstein, and
  Justin~M Solomon.
\newblock Dynamic graph cnn for learning on point clouds.
\newblock {\em Acm Transactions On Graphics (tog)}, 38(5):1--12, 2019.

\bibitem{xie2019pix2vox}
Haozhe Xie, Hongxun Yao, Xiaoshuai Sun, Shangchen Zhou, and Shengping Zhang.
\newblock Pix2vox: Context-aware 3d reconstruction from single and multi-view
  images.
\newblock In {\em Proceedings of the IEEE/CVF International Conference on
  Computer Vision}, pages 2690--2698, 2019.

\bibitem{yang2020robust}
Bo Yang, Sen Wang, Andrew Markham, and Niki Trigoni.
\newblock Robust attentional aggregation of deep feature sets for multi-view 3d
  reconstruction.
\newblock {\em International Journal of Computer Vision}, 128(1):53--73, 2020.

\bibitem{yang2008stereo}
Qingxiong Yang, Liang Wang, Ruigang Yang, Henrik Stew{\'e}nius, and David
  Nist{\'e}r.
\newblock Stereo matching with color-weighted correlation, hierarchical belief
  propagation, and occlusion handling.
\newblock {\em IEEE Transactions on Pattern Analysis and Machine Intelligence},
  31(3):492--504, 2008.

\bibitem{yang2021mvs2d}
Zhenpei Yang, Zhile Ren, Qi Shan, and Qixing Huang.
\newblock Mvs2d: Efficient multi-view stereo via attention-driven 2d
  convolutions.
\newblock {\em arXiv preprint arXiv:2104.13325}, 2021.

\bibitem{yao2018mvsnet}
Yao Yao, Zixin Luo, Shiwei Li, Tian Fang, and Long Quan.
\newblock Mvsnet: Depth inference for unstructured multi-view stereo.
\newblock In {\em Proceedings of the European Conference on Computer Vision
  (ECCV)}, pages 767--783, 2018.

\bibitem{yariv2020multiview}
Lior Yariv, Yoni Kasten, Dror Moran, Meirav Galun, Matan Atzmon, Basri Ronen,
  and Yaron Lipman.
\newblock Multiview neural surface reconstruction by disentangling geometry and
  appearance.
\newblock {\em Advances in Neural Information Processing Systems}, 33, 2020.

\bibitem{yu2021attention}
Anzhu Yu, Wenyue Guo, Bing Liu, Xin Chen, Xin Wang, Xuefeng Cao, and Bingchuan
  Jiang.
\newblock Attention aware cost volume pyramid based multi-view stereo network
  for 3d reconstruction.
\newblock {\em ISPRS Journal of Photogrammetry and Remote Sensing},
  175:448--460, 2021.

\bibitem{yu2020fmvs}
Zehao Yu and Shenghua Gao.
\newblock {Fast-MVSNet}: Sparse-to-dense multi-view stereo with learned
  propagation and gauss-newton refinement.
\newblock In {\em Conference on Computer Vision and Pattern Recognition
  (CVPR)}, 2020.

\bibitem{yuan2021vanet}
Yi Yuan, Jilin Tang, and Zhengxia Zou.
\newblock Vanet: a view attention guided network for 3d reconstruction from
  single and multi-view images.
\newblock In {\em 2021 IEEE International Conference on Multimedia and Expo
  (ICME)}, pages 1--6. IEEE, 2021.

\bibitem{zheng2014patchmatch}
Enliang Zheng, Enrique Dunn, Vladimir Jojic, and Jan-Michael Frahm.
\newblock Patchmatch based joint view selection and depthmap estimation.
\newblock In {\em Proceedings of the IEEE Conference on Computer Vision and
  Pattern Recognition}, pages 1510--1517, 2014.

\bibitem{Zhou2018o3d}
Qian-Yi Zhou, Jaesik Park, and Vladlen Koltun.
\newblock {Open3D}: {A} modern library for {3D} data processing.
\newblock {\em arXiv:1801.09847}, 2018.

\bibitem{zitnick2000cooperative}
C~Lawrence Zitnick and Takeo Kanade.
\newblock A cooperative algorithm for stereo matching and occlusion detection.
\newblock {\em IEEE Transactions on pattern analysis and machine intelligence},
  22(7):675--684, 2000.

\end{thebibliography}
}

\end{document}